\documentclass[10pt,twocolumn,letterpaper]{article}

\usepackage[pagenumbers]{wacv} 

\usepackage{graphicx}
\usepackage{amsmath}
\usepackage{amssymb}
\usepackage{booktabs}
\usepackage{times}
\usepackage{epsfig}
\usepackage{graphicx}
\usepackage{amsmath}
\usepackage{amssymb}
\usepackage{xcolor}  
\usepackage{multirow}
\usepackage{array}
\usepackage{bbm}
\usepackage[normalem]{ulem}
\usepackage{xpatch}

\usepackage{booktabs}
\usepackage[noend]{algorithmic}
\usepackage{algorithm}
\newcolumntype{M}[1]{>{\centering\arraybackslash}m{#1}}

\usepackage[pagebackref,breaklinks,colorlinks]{hyperref}

\usepackage[capitalize]{cleveref}
\crefname{section}{Sec.}{Secs.}
\Crefname{section}{Section}{Sections}
\Crefname{table}{Table}{Tables}
\crefname{table}{Tab.}{Tabs.}

\def\wacvPaperID{147} 
\def\confName{WACV}
\def\confYear{2024}

\begin{document}

\title{Watch Where You Head: A View-biased Domain Gap in Gait Recognition and Unsupervised Adaptation}

\author{Gavriel Habib\thanks{\;\;\; The authors contributed equally}$^{*}$
\and
Noa Barzilay$^{*}$
\and
Or Shimshi
\and 
Rami Ben-Ari
\and
Nir Darshan
\and
\mbox{}
\centerline{OriginAI, Israel}
\\
{\tt\small\{gavrielh, noab, ors, ramib, nir\}@originai.co}}

\maketitle

\begin{abstract}
Gait Recognition is a computer vision task aiming to identify people by their walking patterns. Although existing methods often show high performance on specific datasets, they lack the ability to generalize to unseen scenarios. Unsupervised Domain Adaptation (UDA) tries to adapt a model, pre-trained in a supervised manner on a source domain, to an unlabelled target domain. There are only a few works on UDA for gait recognition proposing solutions to limited scenarios. In this paper, we reveal a fundamental phenomenon in adaptation of gait recognition models, caused by the bias in the target domain to viewing angle or walking direction. We then suggest a remedy to reduce this bias with a novel triplet selection strategy combined with curriculum learning. To this end, we present Gait Orientation-based method for Unsupervised Domain Adaptation (GOUDA). We provide extensive experiments on four widely-used gait datasets, CASIA-B, OU-MVLP, GREW, and Gait3D, and on three backbones, GaitSet, GaitPart, and GaitGL, justifying the view bias and showing the superiority of our proposed method over prior UDA works.
\end{abstract}

\begin{figure*}[t!]
    \centering
    \begin{subfigure}[t]{0.25\textwidth}
        \centering
        \includegraphics[height=1.3in]{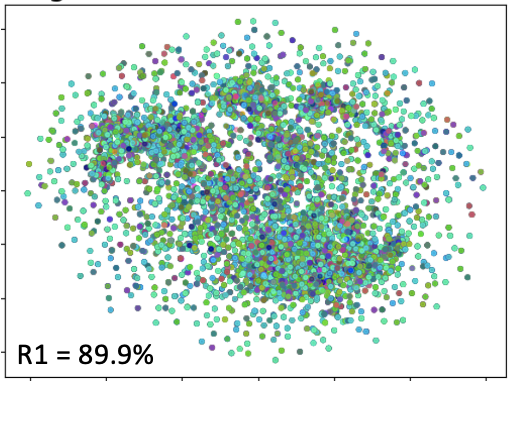}
        \caption{Target Domain = Source Domain}
    \label{fig:view_distribution_source}
    \end{subfigure}
    ~ 
    \begin{subfigure}[t]{0.25\textwidth}
        \centering
        \includegraphics[height=1.3in]{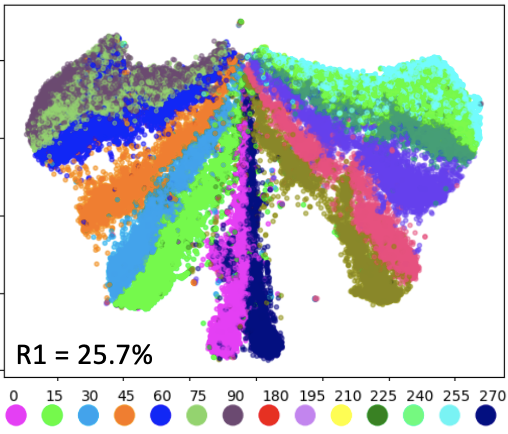}
        \caption{Direct Testing}
    \label{fig:view_distribution_cross}
    \end{subfigure}
    ~ 
    \begin{subfigure}[t]{0.25\textwidth}
        \centering
        \includegraphics[height=1.3in]{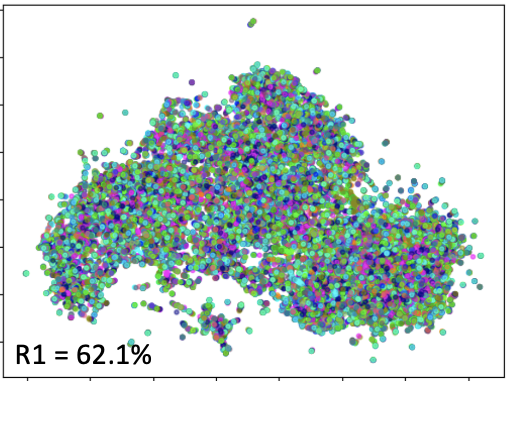}
        \caption{Direct Testing + GOUDA}
    \label{fig:view_distribution_gouda}
    \end{subfigure}
    \caption{Gait embedding visualization (UMAP~\cite{mcinnes2018umap}) for domain transfer between GREW as source and OU-MVLP as the target domain. Points are color coded by viewing angle.  Fig.~\ref{fig:view_distribution_source} showcases the single domain scenario, in which source and target domains are the same (OU-MVLP). Fig.~\ref{fig:view_distribution_cross} presents the direct testing scenario, in which the model was trained on GREW and tested on OU-MVLP. The Rank-1 performance is dropped and a strong viewing angle bias is observed. Fig.~\ref{fig:view_distribution_gouda} presents the direct testing scenario after applying GOUDA using the viewing angles provided in the dataset (just for sake of illustration). Our method is capable of disentangling the viewing angle bias to improve the recognition performance, showing a relatively uniform view distribution similar to the source domain.}
\label{fig:view_distribution}
\end{figure*}

\section{Introduction}
\label{sec:introduction}
Gait recognition is a biometric method that analyzes and identifies individuals based on their walking patterns. It involves the analysis of various characteristics such as stride, body shape, motion postures, and limb movements. Gait recognition has been widely used in different application areas such as sport~\cite{zhang2021gait}, health monitoring~\cite{sun2020gait}, forensic analysis~\cite{macoveciuc2019forensic} as well as security and surveillance~\cite{ran2010applications}. However, compared to different biometrics, like face and fingerprint recognition, gait is a challenging task due to complex conditions such as clothes changing, object carrying, and diverse camera views. 
Existing methods achieve high performance on specific datasets~\cite{chao2019gaitset, fan2020gaitpart, lin2021gait}. GaitSet~\cite{chao2019gaitset} learns from a set of gait silhouettes, GaitPart~\cite{fan2020gaitpart} proposes body part-based model, and GaitGL~\cite{lin2021gait} utilizes global and local visual information. However, the ability of these methods to generalize to new datasets is limited due to domain gaps. Domain gaps can be caused by \eg different camera settings or clothing conditions (changed between seasons), making the gait sequences appear differently across use cases.

Unsupervised Domain Adaptation (UDA) aims to transfer a model learned from a {\it source} domain with labeled data to an unseen {\it target} domain without labels. UDA is common in different computer vision tasks, such as object classification~\cite{zhu2023patch, xu2021cdtrans}, semantic segmentation~\cite{chen2022pipa, hoyer2022hrda}, and person re-identification~\cite{wang2022uncertainty, li2021adadc, bertocco2021unsupervised}. 

UDA for Gait Recognition addresses the case in which the source and target domains include different identities. One main concern for gait recognition task is the ability to recognize an individual over different camera views and walking directions. The interplay between camera positioning and an individual's walking direction within the 3D realm defines what is referred to as the {\it viewing angle} in literature~\cite{fan2023opengait, takemura2018multi, yu2006framework}. Gait recognition models are trained to generalize over different viewing angles, particularly tested by robustness on such scenarios~\cite{takemura2018multi, yu2006framework}.

In this paper, we consider a case where a trained model on a labeled source domain should be adapted to a target domain without any access to the target labels or the source data (that could be unavailable due to privacy issues). To this end, we first use the source model to generate an initial representation of the target domain followed by an adaptation scheme to adapt the pre-trained backbone to the new domain. Such adaptation is often carried out by applying a contrastive loss such as triplet loss~\cite{hermans2017defense}. Without access to the labels in the target domain, prior works used different cues to distinguish (approximately) between the samples~\cite{zheng2021trand, wang2022indoor}. In this paper, we propose GOUDA - Gait Orientation-based method for Unsupervised Domain Adaptation. Our method is driven mainly by a strong observation of the gait view distribution in the target domain (we consider yaw angle here). Fig. \ref{fig:view_distribution} shows an example of view distribution on OU-MVLP dataset~\cite{takemura2018multi}. The depicted UMAPs show the distribution of gait representations color coded by corresponding viewing angles. While in the source domain (Fig.~\ref{fig:view_distribution_source}) the distribution is nearly uniform indicating that relations in the embedding space are {\it invariant} to the view, the clustered colors in the target domain (Fig.~\ref{fig:view_distribution_cross}) imply a strong bias toward the viewing angle. In the target domain, nearby points are likely to be from similar view and not necessarily the same identity (see Rank-1 accuracies in the figure). This view bias causes a significant drop in recognition. Fig.~\ref{fig:view_distribution_gouda} shows that GOUDA is capable of disentangling the view bias, modifying the embedding space in the target domain to improve the recognition accuracy. We show that this fundamental observation and correction method is a key contributor to gait UDA methods as evaluated over various gait backbones and cross domain datasets. Essentially, we suggest a novel triplet selection strategy, based on this key observation. Our method also uses a self-supervised method and curriculum learning, to further boost the results. To summarize, these are our key contributions:


\begin{itemize}
\item We reveal a fundamental bias in cross-domain gait recognition models toward viewing angles.
\item We propose GOUDA, a UDA method with a novel view-based Triplet Selection strategy, mitigating the observed viewing angle bias and leading to a significant improvement in gait UDA task.
\item We conduct extensive experiments with three backbones and with 14 source and target combinations, outperforming prior state-of-the-art by a large gap.
\end{itemize}

\section{Related Work}
\label{sec:related_work}
Existing gait recognition methods can be grouped into two primary categories: model-based approaches and appearance-based approaches.

Model-based approaches~\cite{li2020end, teepe2021gaitgraph, LIAO2020107069, xue2022spatial} focus on capturing the underlying structure and kinematic features of the human body to learn the walking patterns for recognition. These methods utilize structural models, such as 2D or 3D skeleton keypoints~\cite{li2020end, teepe2021gaitgraph, LIAO2020107069, zhu2023gait} as their input representation, aiming to achieve robustness to clothing and object carrying conditions. However, they often face challenges in estimating body parameters in low-resolution videos. Appearance-based methods extract gait representations directly from RGB images or binary silhouette sequences without explicitly modeling the human body. In this context, Silhouette-based recognition is widely used in the literature e.g., template-based methods \cite{wang2011human, li2020gait,han2005individual,wang2011human} that treat entire gait sequences as templates, sequence-based methods that consider gait sequences as videos~\cite{liang2022gaitedge, Huang_2021_ICCV, lin2021gait, lin2020gait, huang2021context, fan2022learning} or as an unordered image set~\cite{chao2019gaitset, hou2020gait}. In our study, we use appearance-based gait recognition models, specifically employing silhouette sequences or unordered sets as input data.

\subsection{Unsupervised Domain Adaptation (UDA)}
In general UDA task~\cite{wilson2020survey, liu2022deep}, domain-invariant feature learning aims to align the source and target distributions by minimizing divergence measures~\cite{li2019cross} or by utilizing adversarial training~\cite{ganin2016domain, Tzeng_2017_CVPR}. Other UDA approaches employ ensemble methods~\cite{french2017self} or adapt batch normalization statistics from source to target domain~\cite{li2018adaptive}. A different approach applies clustering using pseudo-labels~\cite{wang2022uncertainty, li2021adadc, bertocco2021unsupervised}, in the target domain. Our approach draws inspiration from aligning the source and target distributions.

UDA has been employed in the re-identification task coping with variations in view, camera settings, and environmental conditions, over different domains. In this respect, UNRN~\cite{zheng2021exploiting} is a clustering-based method with uncertainty-guided optimization driven by a teacher-student framework. SECRET~\cite{he2022secret} utilizes global and local features for a self-consistent pseudo-label refinement. A triplet strategy was used by Bertocco et al.~\cite{bertocco2021unsupervised} based on the diversity of cameras within a cluster. However, these studies address a different task of re-identification. In contrast to the person re-identification task, gait recognition focuses on videos of binary silhouettes or human skeletons rather than RGB images, to be invariant to the RGB appearance information. 
 
\begin{figure}
\centerline{\includegraphics[width=0.8\linewidth]{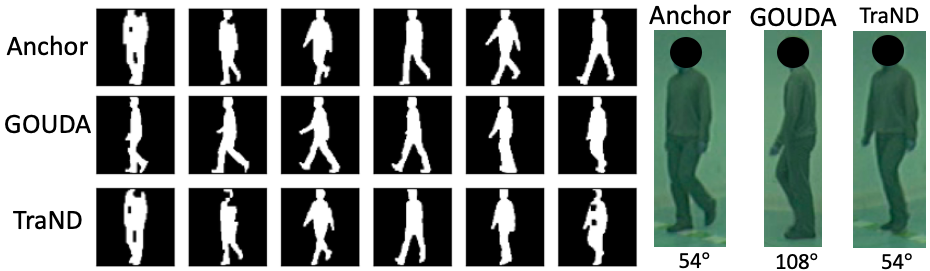}}
\caption{An example of positive sample selection. We use the setting with GaitSet and OU-MVLP (source) to CASIA-B (target). TraND~\cite{zheng2021trand} usually selects positive samples with the exact same view as the anchors due to its strategy to bring adjacent samples closer. Contrary to that, GOUDA selects positives with different views by default. Here, the anchor view is 54$^{\circ}$, and the positive sample selected by GOUDA is of view 108$^{\circ}$. In both cases, the positives are correct identity-wise. By bringing closer samples of different views, our method improves the cross-view benchmarks and outperforms TraND, as shown in Table~\ref{table:SOTA_compare}.}
\label{fig:positive_selection_example}
\end{figure}

Despite the necessity of UDA in Gait Recognition, only a few studies have addressed this specific task~\cite{zheng2021trand, wang2022indoor}. Both of these methods attempt for a fine-tuning adaptation on the target domain. From Indoor to Outdoor (FI2O)~\cite{wang2022indoor} conducts an adaptation step by using a pseudo-labeling strategy in the target domain, to adapt a model trained on an indoor environment to an outdoor using noisy labels. Their method relies on a clustering approach that assumes the existence of multiple sequences per person in the target dataset, which may not always be available.
TraND~\cite{zheng2021trand} attempts to close the domain gap by a neighborhood selection strategy (for training) that relies on an entropy-based confidence measure. Effectively, its main effort is to keep bringing closer adjacent samples, with high confidence of belonging to the same identity. As a result, TraND tends to favor positive samples with similar views. Due to this bias, TraND struggles to deal with the common case of different views. In contrast to TraND, GOUDA adopts a different strategy by selecting positive samples from diverse views (see Fig.~\ref{fig:positive_selection_example}). It enriches the training set and contributes to the development of a model that is robust against variations in views. We also show that adding self-supervised learning on top of our triplet selection method, boosts the results, reducing the gap toward the upper bound of fine-tuning with labels.

\section{Method}
\label{sec:method}
\begin{figure}
\centerline{\includegraphics[width=0.9\linewidth]{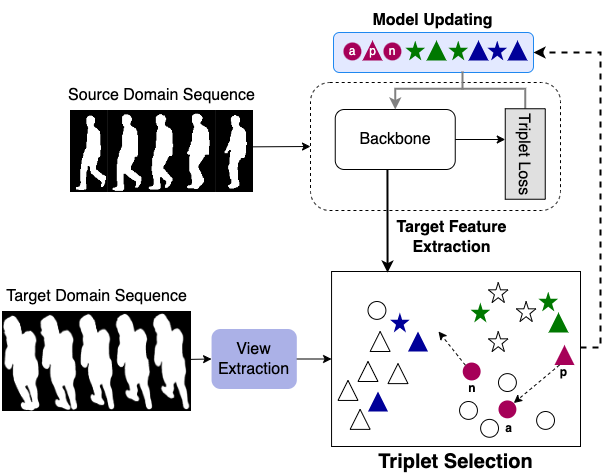}}
\caption{An overview of GOUDA. First, a pre-trained backbone, trained on the source domain, is used to extract gait features of sequences from an unseen target domain. Each sequence in the figure is characterized by a distinct shape, representing its corresponding view extracted using View Extraction module. Second, the Triplet Selection algorithm is applied, aiming to push away sequences of similar views and bring closer different ones in the embedding space. The colored shapes depict the selected triplets. The triplets are then used to update the backbone using triplet loss.}
\label{fig:Overview}
\end{figure}

Figure~\ref{fig:Overview} presents an overview of our method with three main stages. First, a gait model, pre-trained on a source domain, is used as a feature extractor for an unseen target domain. In the embedding space, each sequence is described by its evaluated viewing angle, while its label (identity) is unknown. Second, our Triplet Selection approach is applied, aiming to push away sequences of similar views while bringing closer different views. In order to increase the probability that the triplets are identity-wise correct, we further adopt a Curriculum Learning framework~\cite{bengio2009curriculum} based on a certainty measure. Lastly, the selected triplets are used to fine-tune the network using standard triplet loss.

\subsection{View Extraction}
\label{sec:view_extraction}
In gait recognition, the view is defined as the walking direction of the person relative to the camera. To estimate the view of a gait sequence in the target domain, we use the provided 2D keypoints (representing the posture of human body parts). First, we apply MHFormer~\cite{li2022mhformer} to regress the 3D keypoints per frame. Then, we estimate the sequence view (yaw angle) using the median 3D keypoints of relevant body parts - hips and shoulders. For real application uses, the 2D keypoints can be estimated from RGB images.

\subsection{Triplet Selection}
\label{sec:triplet_selection}

This section provides a description of the Triplet Selection algorithm, which aims to improve performance on the target domain by disentangling the view bias. Ideally, we aim to find correct triplets such that, given an anchor, the positive sample has the same identity but a different view, and the negative sample belongs to a different identity but of similar view.

{\bf Negative Sample Selection. }Based on the observation that gait recognition models strongly emphasize view-based features in the target domain (Fig.~\ref{fig:view_distribution}), we assume that within a neighborhood of samples with a similar view, there are likely to exist samples of different identities. 

{\bf Positive Sample Selection. } Gait models demonstrate a certain level of learned general gait features. Hence, we assume that among all samples with different views, the ones with the same identities are reasonable to be the closest\footnote{In the supplementary materials, we present further intuition and justification, including a shared visualization of views and identities.}. In Fig.~\ref{fig:view_distribution_cross}, it is implied by the cross-view Rank-1 accuracy (25.7\%), which is much higher than random (±0.02\%).

In the embedding space, we prioritize pushing away similar-view samples and bringing closer together cross-view samples. We use various techniques, such as curriculum learning, to increase the identity-wise certainty.

The input to the algorithm (pseudo-code presented in Alg.~\ref{algo:triplet_selection}) is an unlabeled target domain set ${S}$ := {$\{ (x_{r}, v_{r})\}_{r=1}^R$}, where $R$ is the number of gait sequences, $x_{r} \in \mathbb{R}^{W \times H \times T}$ is the $r$-th sequence with $T$ frames of size $W \times H$, and $v_{r} \in \mathbb{R}$ is its estimated view. Each gait sequence $x_{r}$ is represented by its embedding $e_{r}$=$f(x_{r})$, where $f(\cdot)$ is a gait model pre-trained in a supervised manner on the source domain. In addition, we define the following hyper-parameters: margin $m$, and view thresholds $T_{s}, T_{c}$ such that $T_{s} \leq T_{c}$. We use $v_j$ as an indicator (in the range of $[0, 360)$), for separation of the data for each anchor $a$ into two mutually exclusive sets. Considering an anchor $a$, the sample $j$ belongs to the set $\chi^{a}_{similar}$, if $\vert v_{a}-v_{j}\vert < T_{s}$, and belongs to $\chi^a_{cross}$, if $\vert v_{a}-v_{j}\vert > T_{c}$. The triplets $\{a, p, n\}$ are selected such that $n \in \chi^{a}_{similar}$ and $p \in \chi^a_{cross}$.

\begin{figure}
\centerline{\includegraphics[width=0.45\linewidth]{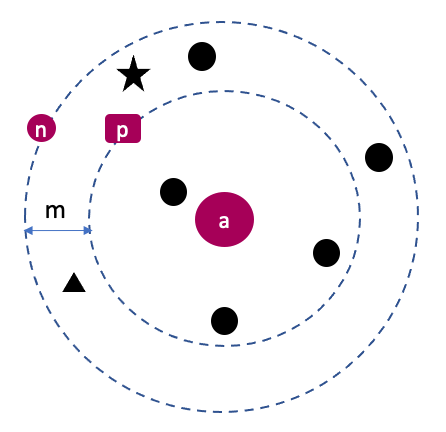}}
\caption{Triplet selection illustration (shape describes view). Given an anchor $a$, the positive $p$ is the closest sample with a different view. The negative $n$ is the farthest sample (up to margin $m$ from $p$) with a similar view as $a$. A triplet $\{a, p, n\}$ is ``valid" if exists at least one similar-view sample closer to $a$ than $n$.}
\label{fig:TripletSelection}
\end{figure}

Given a set of embeddings and views {$\{ (e_{r}, v_{r})\}_{r=1}^R$}, we now calculate the distances matrix $D \in 	\mathbb{R}^{R \times R}$, such that $D_{i,j}:= d(e_i, e_j)$, where $d$ is the cosine distance metric. In order to increase the certainty of triplets to be identity-wise correct, we select them as follows. The positive $p$ is defined as the closest sample to $a$ in $\chi^a_{cross}$ (line~\ref{eq_positive} in Alg.~\ref{algo:triplet_selection}). We choose $n \in \chi^{a}_{similar}$ to be the farthest sample from $a$ within a fixed margin $m$ from $p$ (line~\ref{eq_negative} in Alg.~\ref{algo:triplet_selection}). Eventually, the triplet $\{a, p, n\}$ is ``valid" if exists at least one sample in $\chi^{a}_{similar}$ that is closer to $a$ than $n$ (line~\ref{eq_valid_triplet_if} in Alg.~\ref{algo:triplet_selection}). Both restrictions on $n$ are used to increase the probability that $a$ and $n$ do not share the same identity, assuming that the closest similar-view sample is more likely to share the same identity. The margin $m$ is incorporated into the negative sample selection (line~\ref{eq_negative} in Alg.~\ref{algo:triplet_selection}) in order to align with the objective of triplet loss. An illustration of the Triplet Selection algorithm is shown in Fig.~\ref{fig:TripletSelection}, and described in Algorithm~\ref{algo:triplet_selection}. The algorithm is applied iteratively during the training process (Section~\ref{sec:curriculum_learning}).

\begin{algorithm}[t]
\caption{Triplet Selection}
\label{algo:triplet_selection}
\begin{algorithmic}[1]
 \STATE \textbf{function} \textsc{Triplet Selection}\{$D$, $\{v_{r}\}_{r=1}^R$\}
 \FOR{$a \gets 1$ to $R$}
  \STATE{$\chi^a_{similar} \leftarrow \{ j \mid \vert v_{a}-v_{j}\vert < T_{s}, a \neq j \}$} \label{eq_x_sim}
  \STATE{$\chi^a_{cross} \leftarrow \{ j \mid \vert v_{a}-v_{j}\vert > T_{c}\}$} \label{eq_x_cross}
  \STATE{$p \leftarrow \chi^a_{cross} [\underset{j\in \chi^a_{cross}}{\arg\min} \{ D_{a,j}$ \}]} \label{eq_positive}
    \STATE{$n \leftarrow \chi^a_{similar} [\underset{j  \in \chi^a_{similar}}{\arg\max} \{ D_{a,j} \mid D_{a,j} < D_{a,p}+m \}]$} \label{eq_negative}
    \STATE{\textbf{if } {$\exists j \in \chi^a_{similar}$ \text{ s.t: } $D_{a,j} < D_{a,n}$} \textbf{ then}}  \label{eq_valid_triplet_if}\\
     \STATE {\quad\quad $\mathcal{V}$: Valid Triplets $\gets \{a,p,n\}$} \label{eq_valid_triplet_set}
    \STATE {\quad\quad $\mathcal{C}$: Confidence Scores $\gets 1-D_{a,p}$} \label{eq_conf_score_set}
 \ENDFOR
 \STATE {Selected Triplets $\gets \mathcal{V}[\underset{\text{top $q$\%}} {\arg\max  \mathcal{C}}]$}
 \RETURN{Selected Triplets}
\end{algorithmic}
\end{algorithm}

\subsection{Curriculum Learning}
\label{sec:curriculum_learning}
Inspired by~\cite{bengio2009curriculum, zheng2021trand}, we apply the Triplet Selection algorithm in four different stages during training, in a from-easy-to-hard manner. We define a \textit{triplet confidence score} as the cosine similarity between the anchor and the positive samples ($1 - D_{a,p}$). We select the top $q\%$ valid triplets $\mathcal{V}$ according to their confidence scores $\mathcal{C}$, while $q$ gets higher in each stage. We keep improving target domain performance by starting from updating the model with a small amount but high-quality data and carefully extending the dataset size.

\subsection{Stopping Criteria}
\label{sec:stopping_criteria}
In contrast to previous works~\cite{wang2022indoor, zheng2021trand}, we introduce an indicative stopping criteria (SC) using a validation set. Although training with a fixed number of iterations is simple, it lacks control over the distribution obtained in the embedding space, due to an over-fitting risk. Our SC is driven by the desire to achieve greater diversity in view within the neighborhoods. Given a validation set of size $N_{val}$, for each sample $a$ we compute its $K$ nearest neighbors, $KNN(a)$, using the cosine distance metric. Then, we count how many of the $K$ samples are of similar view as $a$. The final $SC$ value is the average of all counts (Eq.~\ref{eq:stopping_criteria}). Eventually, we choose the checkpoint in which $SC$ is minimal.
\label{sec:stopping_criteria}
\begin{align}
SC = & \frac{1}{{N_{val}}} \sum_{a=1}^{N_{val}} \sum_{i \in KNN(a)}  \mathbbm{1}{ [ i \in \chi^a_{similar} ] } \longrightarrow min
\label{eq:stopping_criteria}
\end{align}

\subsection{Model Updating}
\label{sec:model_updating}
In the adaptation stage,  we optimize the backbone $f$ with two triplet loss objectives: GOUDA and self-supervision loss. Suppose $\{a_i, p_i, n_i\}$ is a selected triplet, GOUDA loss is given by:
\label{sec:GOUDA_loss}
\begin{align}
\mathcal{L}_{GOUDA} & = \sum_{i=1}^{N}{[D_{a_i,p_i} - D_{a_i,n_i} + m]_{+}}
\end{align}
where $N$ represents the number of triplets, $m$ is the triplet loss margin and $[x]_{+}$ is max($x$, 0). Note that $a$ and $n$ share a similar-view, different from $p$. Our self-supervision loss (SSL) is defined by:
\begin{align}
\mathcal{L}_{SSL} & = \sum_{i=1}^{N}{[D_{a_i,\Tilde{a}_i} - D_{a_i,n_i} + m]_{+}}
\label{eq:SSL_losses}
\end{align}
In the self-supervision loss we replace $p$ with $\Tilde{a}$, an augmented version of $a$. The augmentations are two random sub-sequences of the same gait sequence. Eventually, the losses are linearly combined to our domain adaptation loss:
\begin{align}
\mathcal{L} & = \mathcal{L}_{GOUDA} + \mathcal{L}_{SSL}
\label{eq:model_losses}
\end{align}
The selected triplets are randomly extracted for training.

\section{Experiments}
\label{sec:experiments}

\subsection{Datasets}
\label{sec:datasets}
In this paper we evaluate the proposed GOUDA on four well-known gait datasets, CASIA-B~\cite{yu2006framework}, OU-MVLP~\cite{takemura2018multi}, Gait3D~\cite{Zheng_2022_CVPR}, and GREW \cite{Zhu_2021_ICCV}.

\textbf{CASIA-B}~\cite{yu2006framework} is a popular indoor gait dataset. It contains 13,640 gait sequences, including 124 subjects with 11 views. Each subjects' videos include 3 different settings, normal walking (NM), walking with a bag (BG), and walking with a coat (CL). The first 74 subjects are used for the training set and the rest 50 subjects are used for the test set.

\textbf{OU-MVLP}~\cite{takemura2018multi, an2020performance} is a large indoor gait dataset. It contains 10,307 subjects with 14 views. According to the provided protocol~\cite{chao2019gaitset}, 5,153 subjects are used for the training set, and the remaining 5,154 subjects are used for testing.

\textbf{Gait3D}~\cite{Zheng_2022_CVPR} is a 3D representation-based gait dataset captured in the supermarket. The dataset contains 4,000 subjects and over 25,000 sequences extracted from 39 cameras. The training set includes 3,000 subjects, and the test set includes the remaining 1,000 subjects.

\textbf{GREW}~\cite{Zhu_2021_ICCV} is a large-scale in-the-wild gait recognition dataset. The dataset contains 26,345 subjects and 128,671 sequences extracted from 882 cameras. 
We adopt 2 protocols to evaluate the dataset~\cite{Zhu_2021_ICCV},~\cite{wang2022indoor}. According to the first protocol~\cite{Zhu_2021_ICCV}, the training set contains 20,000 subjects and the test set includes 6,000 subjects. The second protocol~\cite{wang2022indoor} splits the original training set, the first 2,000 subjects are utilized for training and the last 2,000 subjects are utilized for test. We refer to the latter partition as GREW$^\ast$. 
 
\subsection{Implementation details}
\label{sec:implementation_details}
We use three different well-known gait backbones - GaitSet~\cite{chao2019gaitset}, GaitPart~\cite{fan2020gaitpart}, and GaitGL~\cite{lin2021gait} to evaluate the proposed GOUDA. The input silhouettes in all datasets are resized to 64$\times$44. Each backbone was first trained on the source domain data, using the implementations provided in OpenGait code~\cite{fan2023opengait}. The training set in each target domain was partitioned into separate training and validation sets. Specifically, we created the validation set by selecting the last 10\% of subjects from the training set. 

In the model updating stage, all backbones were trained using the loss function as described in Eq. \ref{eq:model_losses}, with $m=0.2$. View thresholds $T_{s}$, and $T_{c}$ are set to $10^\circ$ and $20^\circ$ respectively. We adopt Adam optimizer~\cite{kingma2014adam} with a learning rate of 1e-5 and weight decay of 5e-4. In every training iteration, a batch size of 32$\times$4 was employed, where the number 4 corresponds to the inclusion of two different augmentations applied to the anchor sample ($a, \Tilde{a}$), a positive sample ($p$), and a negative sample ($n$). The final weights of the model were selected according to the optimal results of the stopping criteria (cf. Sec. \ref{sec:stopping_criteria})  on the validation set. $K$ value of the SC was set to 5. The curriculum learning strategy includes four stages, where each stage involves selecting a higher percentage of top $q\%$ valid triplets: $10$, $25$, $50$, and $100$\footnote{Ablation study on $q$ values is shown in the Supplementary Materials.}. The amount of training iterations after each stage is proportional to the number of selected triplets. Specifically, we move to the next stage after each selected triplet appears in training 10 times on average.
 
\section{Evaluation}
\label{sec:evaluation}

\subsection{Results}
\label{sec:results}

\begin{table*}
\renewcommand{\arraystretch}{1.1}
  \begin{center}
    {\small{
\begin{tabular}{ M{1.3cm}| M{1.2cm} | M{0.7cm} M{0.7cm}  M{0.7cm} M{0.7cm}  M{0.7cm} M{0.7cm} | M{0.7cm} M{0.7cm} | M{0.7cm} M{0.7cm} | M{0.7cm} M{0.7cm} }
\toprule[1.2pt]
\multirow{4}{*}{\shortstack{Source \\ Dataset}} & \multirow{4}{*}{Backbone} & \multicolumn{12}{c}{Target Dataset} \\
\cline{3-14}
& & \multicolumn{6}{c|}{CASIA-B} & \multicolumn{2}{c|}{\multirow{2}{*}{OU-MVLP}} & \multicolumn{2}{c|}{\multirow{2}{*}{GREW}} & \multicolumn{2}{c}{\multirow{2}{*}{Gait3D}} \\
& & \multicolumn{2}{c}{NM} & \multicolumn{2}{c}{BG} & \multicolumn{2}{c|}{CL} & & & & & & \\
\cline{3-14}
& & GO & DT & GO & DT & GO & DT & GO & DT & GO & DT & GO & DT \\
\midrule[1.2pt]
\multirow{2}{*}{CASIA-B} & \shortstack{GaitSet \\ GaitPart  \\ GaitGL} &  & & \multicolumn{2}{c }{-}  & &  & \shortstack{\textbf{27.9} \\ \textbf{27.5} \\ \textbf{34.0}} 
&  \shortstack{9.6 \\ 10.8\\ 16.2} & \shortstack{9.4 \\ \textbf{15.3} \\ \textbf{14.6}} & \shortstack{9.4 \\ 9.3 \\ 11.6} & \shortstack{\textbf{9.3} \\ \textbf{11.5}  \\ \textbf{10.9} } & \shortstack{7.7 \\ 6.6 \\ 7.6} \\
\midrule
\multirow{2}{*}{OU-MVLP} & \shortstack{GaitSet \\ GaitPart  \\ GaitGL} & \shortstack{\textbf{87.0} \\ \textbf{91.5}  \\ \textbf{92.8}} & \shortstack{74.0 \\ 73.9 \\ 81.7} & \shortstack{\textbf{68.0} \\ \textbf{78.4} \\ \textbf{80.9}} & \shortstack{55.5 \\56.9 \\ 71.5} & \shortstack{\textbf{27.2} \\ \textbf{38.7} \\ \textbf{44.6}} & \shortstack{16.4 \\20.7 \\28.8} &\multicolumn{2}{c|}{-} & \shortstack{11.7 \\ \textbf{21.0} \\ \textbf{25.7}} & \shortstack{\textbf{12.3}\\ 14.8\\ 21.4} & \shortstack{\textbf{13.3} \\ \textbf{18.8}\\ \textbf{21.6}} & \shortstack{13.0 \\ 11.2 \\ 16.0} \\
\midrule
\multirow{2}{*}{GREW} & \shortstack{GaitSet \\ GaitPart  \\ GaitGL} & \shortstack{\textbf{76.9} \\ \textbf{81.4}  \\ \textbf{82.7}} & \shortstack{65.6 \\ 69.2 \\ 69.8} & \shortstack{\textbf{56.8} \\ \textbf{68.3} \\ \textbf{73.2}} & \shortstack{44.9 \\ 52.1 \\ 61.1}
& \shortstack{\textbf{26.0}  \\ \textbf{33.9} \\ \textbf{44.6}} & \shortstack{20.8 \\ 25.4 \\ 31.9} & \shortstack{\textbf{38.8} \\ \textbf{43.6} \\ \textbf{44.2}} & \shortstack{21.8 \\ 23.9 \\ 25.7} &\multicolumn{2}{c|}{-} & \shortstack{16.3 \\ \textbf{19.5} \\ \textbf{18.4}} & \shortstack{\textbf{19.0} \\ 19.3 \\ 15.6} \\
\midrule
\multirow{2}{*}{Gait3D} & \shortstack{GaitSet \\ GaitPart  \\ GaitGL} & \shortstack{\textbf{73.4}\\ \textbf{76.7} \\ \textbf{79.0}} & \shortstack{62.8 \\ 61.8 \\ 63.5} & \shortstack{\textbf{51.9} \\ \textbf{61.1} \\ \textbf{65.4} } & \shortstack{45.8 \\ 47.0 \\ 51.2} & \shortstack{\textbf{20.4}  \\ \textbf{24.5} \\ \textbf{24.9} } & \shortstack{11.9 \\ 13.7 \\ 16.3} & \shortstack{\textbf{41.9} \\ \textbf{32.2}  \\ \textbf{37.0}} & \shortstack{20.8 \\ 19.1 \\ 23.7}  & \shortstack{14.9  \\ \textbf{14.9} \\ 10.7 } & \shortstack{\textbf{19.2} \\ 14.2 \\ \textbf{14.3}} & \multicolumn{2}{c}{-} \\
\bottomrule[1.2pt]
\end{tabular}
}}
\end{center}
\caption{\label{table:GOUDAMainTab}Rank-1 accuracy of GOUDA (referred to as GO) compared to the results of Direct Testing (DT) for various backbones~\cite{chao2019gaitset},~\cite{fan2020gaitpart},~\cite{lin2021gait}. This table describes 12 settings of source and target datasets.}
\end{table*}

\textbf{Comparison to Direct Testing.} Our method does not assume special architectural conditions and therefore can be integrated with different kinds of deep backbones\footnote{In the Supplementary Materials, we show that the domain gap presented in Fig.~\ref{fig:view_distribution} is consistent across different gait backbones.}. We evaluate the performance of GOUDA when combined with various backbones, including GaitSet~\cite{chao2019gaitset}, GaitPart~\cite{fan2020gaitpart}, and GaitGL~\cite{lin2021gait}, as shown in Table~\ref{table:GOUDAMainTab}. These results are then compared to Direct Testing performances, \ie models that are trained solely on the source domain in a supervised manner. It is evident that the direct testing results exhibit a certain level of competence in learning general gait features, which contributes to achieving reasonably decent results in the target domains. Nevertheless, these results are still unsatisfactory. Referring to Table~\ref{table:GOUDAMainTab}, it becomes apparent that our method yields substantial improvements in Rank-1 accuracy across almost all settings. Notably, GOUDA demonstrates improvements across multiple backbones, underscoring the algorithm's robustness independent of any specific architecture. This finding reinforces the observation presented in Section \ref{sec:introduction}. GOUDA demonstrates superior performance compared to Direct Testing results in 49 out of 54 cases. In some cases, our method improves the Rank-1 accuracy by a significant margin, for example from 23.9\% to 43.6\%, when using GaitPart trained on GREW (source) to OU-MVLP (target). In the supplementary materials, we present even higher improvements when using ground truth views provided as meta-data in CASIA-B and OU-MVLP datasets, avoiding the noise produced by the View Extraction module. It may suggest that the potential of GOUDA is higher, if using an improved 3D keypoints estimator.

It is apparent that the Rank-1 improvements achieved by GOUDA are considerably higher when employing CASIA-B or OU-MVLP datasets as the target domain, compared to GREW or Gait3D. This disparity can be attributed to the nature of the datasets themselves. CASIA-B and OU-MVLP are indoor datasets captured in controlled environments, whereas GREW and Gait3D were collected in-the-wild. Consequently, the latter datasets present more challenging conditions for gait recognition, potentially leading to lower gains in performance compared to the former datasets. Furthermore, our method relies on estimating a single view through the entire sequence. However, such an approach does not accurately reflect the conditions encountered in datasets captured in the wild.
In practice, one can decompose the gait sequence into snippets with piece-wise constant views and analyze them separately.

\textbf{Comparison with State-of-The-Art.} We compare GOUDA to state-of-the-art UDA gait recognition methods~\cite{wang2022indoor, zheng2021trand} and present them in Table~\ref{table:SOTA_compare}. To ensure a fair comparison, we employ the same backbone in each comparison, GaitGL when comparing with~\cite{wang2022indoor} and GaitSet when comparing with~\cite{zheng2021trand}. In order to compare to TraND~\cite{zheng2021trand}, we trained their model using the implementations provided by the authors. As a reference, we further report the results of UDA methods for person re-identification, UNRN~\cite{zheng2021exploiting} and SECRET~\cite{he2022secret}, applied by~\cite{wang2022indoor}. In these cases, the entire gait sequence was compressed to a single template (GEI~\cite{han2005individual}) as these models get a single image as an input.

As shown in Table~\ref{table:SOTA_compare}, GOUDA significantly outperforms prior works in 9 out of 10 settings. The difference in results between GOUDA and TraND~\cite{zheng2021trand} can be attributed to the fundamental principle on which the TraND approach relies. This approach is centered around an entropy-first neighbor selection strategy focusing solely on the nearest neighbor sample that is likely to share the same view. We believe that due to this reason, TraND is less robust to different views.

\begin{table}
\renewcommand{\arraystretch}{1.05}
  \centering
  \small
  \begin{tabular}{M{0.95cm}| M{0.7cm} M{0.7cm} M{0.7cm} | M{0.7cm} M{0.7cm} M{0.7cm} }
    \toprule[1.2pt]
      \multirow{2}{*}{Method} & \multicolumn{3}{c|}{CASIA-B $\rightarrow$ GREW$^\ast$} & \multicolumn{3}{c}{OU-MVLP $\rightarrow$ GREW$^\ast$} \\
    \cmidrule{2-7}
     & R1 & R5 & R10 & R1 & R5 & R10 \\
     \midrule[1pt]
    UNRN & 20.2 & 32.1 & 39.0 & 12.7 & 22.6 & 27.9 \\
    SECRET & 12.3 & 19.6 & 24.8 & 20.1 & 30.9 & 37.0 \\
    \hline
     GaitGL & 30.0 & 40.5 & 46.6 & 42.1 & 56.2 & 63.0 \\
     FI2O & \textbf{37.4} & 49.2 & 53.8 & 47.0 & 60.2 & 65.7 \\
     GOUDA & 35.6 & \textbf{49.9} & \textbf{56.0} & \textbf{50.4} & \textbf{65.0} & \textbf{71.0} \\
     \midrule[0.7pt]
     \midrule[0.7pt]
    \multirow{2}{*}{Method} & \multicolumn{3}{c|}{\multirow{2}{*}{CASIA-B $\rightarrow$ OU-MVLP}} & \multicolumn{3}{c}{OU-MVLP $\rightarrow$ CASIA-B} \\
        \cline{5-7}
     & & &  & NM & BG & CL \\ 
    \midrule[1pt]
     GaitSet & \multicolumn{3}{c|}{9.6} & 74.0 & 55.5 & 16.4 \\
     TraND & \multicolumn{3}{c|}{15.8} & 74.7 & 56.1 & 16.2 \\
     GOUDA & \multicolumn{3}{c|}{\textbf{27.9}} & \textbf{87.0} & \textbf{68.0} & \textbf{27.2} \\
    \bottomrule[1.2pt]
  \end{tabular}
\caption{\label{table:SOTA_compare} Comparison with the state-of-the-art UDA gait recognition methods, FI2O~\cite{wang2022indoor} and TraND~\cite{zheng2021trand}, and with UDA methods for person re-identification, UNRN~\cite{zheng2021exploiting} and SECRET~\cite{ he2022secret}. Results are reported by Rank-1 (R1), Rank-5 (R5) and Rank-10 (R10). GOUDA is compared to each method in two different settings. GREW$^\ast$ uses the protocol presented in~\cite{wang2022indoor}. Direct testing and domain adaptation results of the UDA gait methods are shown with relevant backbones (as used in the corresponding papers).}
\end{table}

\subsection{Ablation Study}
\label{sec:ablation_study}

To demonstrate the efficacy of the proposed method and assess the influence of various key components, we carried out an ablation study using GaitGL with OU-MVLP (source) to CASIA-B (target) setting. The experimental results are reported in Table~\ref{table:AblationStudy}.

\textbf{Analysis of Various Key Components.} We trained various models that emphasize the impact of key components in GOUDA (rows 2-8 in Table~\ref{table:AblationStudy}). Row 2 highlights the impact of incorporating $m$ in the Triplet Selection process. In this experiment, we used $m=0$ when selecting the negative samples (line~\ref{eq_negative} in Alg.~\ref{algo:triplet_selection}). Interestingly, the results are even lower than those achieved with Direct Testing. This outcome can be attributed to the fact that excluding the margin in the Triplet Selection algorithm leads to a significantly reduced number of valid triplets available for training. Thus, it may potentially result in an overfitted model, as it lacks the necessary diversity and complexity in the training data. In row 3, we demonstrate the effectiveness of incorporating a curriculum learning strategy. In this training process, the model undergoes four stages of triplet selection as well. However, unlike GOUDA, all valid triplets are selected during training ($q=100\%$ in all four stages). 

{\bf Loss Components. } The models in rows 4-5 were trained without $\mathcal{L}_{SSL}$ and $\mathcal{L}_{GOUDA}$, respectively, \ie all triplets related to each one of these loss terms were excluded from $\mathcal{L}$. We observe the significant impact of these loss terms on the results, especially of $\mathcal{L}_{GOUDA}$, which is the key component of the loss.

{\bf Hyperparameters. } In row 6, we explore a different Stopping Criteria (SC) for GOUDA. Instead of using 5 neighbors ($K=5$), we examine the SC with a single nearest neighbor ($K=1$). In this case, the results are comparable to GOUDA. However, selecting an environment of 5 neighbors might be less prone to noise compared to solely considering the nearest neighbor. Rows 7-8 present experiments with different values of $T_{c}$ and $T_{s}$ (defined in lines \ref{eq_x_sim}, \ref{eq_x_cross} of Alg.~\ref{algo:triplet_selection}), where in each study only one threshold was changed. These results are inferior or comparable to those achieved with the original values.

\textbf{GOUDA vs. Triplet Selection Oracle.} To evaluate the effectiveness of our novel Triplet Selection algorithm, we conducted an experiment comparing GOUDA with a Triplet Selection Oracle (row 9). In this particular experiment, we trained the model using solely the correct triplets from the set of valid triplets mentioned in Algorithm~\ref{algo:triplet_selection}. Specifically, the oracle utilizes the target identity labels $l$ and exclusively involves a valid triplet $\{a, p, n\}$ only if $l_{a}$ = $l_{p}$ $\neq$ $l_{n}$. By comparing the results of the oracle with those of GOUDA, it is evident that despite the ability of our method to generate high-quality triplets, there is still a gap to overcome.

\textbf{Supervised Fine-tuning.} 
In row 10, we present the results of supervised fine-tuning, as an upper bound, where target labels are available.  In this experiment, a standard identity triplet loss~\cite{hermans2017defense} and cross entropy loss~\cite{de2005tutorial} were employed to fine-tune the GaitGL (without adaptation) on the target domain for 50K iterations. It is evident that there is still a large potential remained to achieve fully supervised results, especially in BG and CL scenarios.

\begin{table}
\renewcommand{\arraystretch}{1.1}
  \begin{center}
  {\small{
  \begin{tabular}{M{0.35cm} | M{3.5cm}| M{0.8cm} | M{0.8cm} | M{0.8cm}}
\toprule[1.2pt]
\multirow{2}{*}{\#} & \multirow{2}{*}{Method} & \multicolumn{3}{c}{OU-MVLP $\rightarrow$ CASIA-B} \\
\cline{3-5}
    & & NM & BG & CL \\
    \midrule[1pt]
    1 & GaitGL Direct Testing & 81.7 & 71.5 & 28.8 \\
    \hline
    2 & GOUDA w/o $m$ & 68.2 & 43.6 & 20.3 \\    
    \hline
    3 & GOUDA w/o curriculum learning & 88.2 & 72.0 & 36.8 \\
    \hline
    4 & GOUDA w/o $\mathcal{L}_{SSL}$ & 90.8 & 76.7 & 39.0 \\ 
    \hline
    5 & GOUDA w/o $\mathcal{L}_{GOUDA}$ & 87.5 & 72.9 & 41.3 \\ 
    \hline
    6 & GOUDA w/ SC $K=1$ & 92.1 & 79.6 & 45.9 \\
    \hline
    7 & GOUDA w/ $T_{c}=10^\circ$ & 87.8 & 70.2 & 39.8 \\
    \hline
    8 & GOUDA w/ $T_{s}=20^\circ$ & 92.1 & 80.2 & 46.8 \\
    \hline
    \hline
    9 & Triplet Selection Oracle & 96.3 & 90.3 & 55.6 \\
    \hline
    10 & Supervised fine-tuning & 98.3 & 94.7 & 86.7 \\ 
    \hline
    \hline
    11 & \textbf{GOUDA} & 92.8 & 80.9 & 44.6 \\
\bottomrule[1.2pt]
\end{tabular}
}}
\end{center}
\caption{\label{table:AblationStudy} Ablation 
experiments with domain adaptation from OU-MVLP (source) to CASIA-B (target). Average Rank-1 accuracies across all 11 views are shown, excluding identical-view cases.}
\end{table}

\section{Analysis}
\label{sec:discussion}
In this section, we analyze our method, its consequences, and limitations.

\begin{figure}
\centerline{\includegraphics[width=0.57\linewidth]{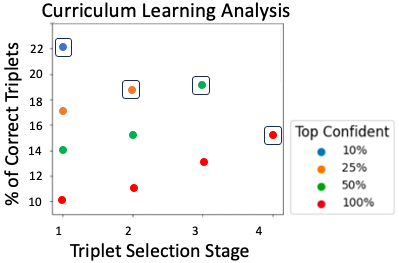}}
\caption{Curriculum Learning Analysis. Triplet Selection is applied $4$ times during training (x-axis). Y-axis shows how many of the selected triplets are correct considering person identities. The colors represent the top confident triplets, and in each stage the chosen percentage is marked by a rectangle (see Section~\ref{sec:curriculum_learning}).}
\label{fig:curriculum_learning_est}
\end{figure}

\begin{figure}
\centerline{\includegraphics[width=0.57\linewidth]{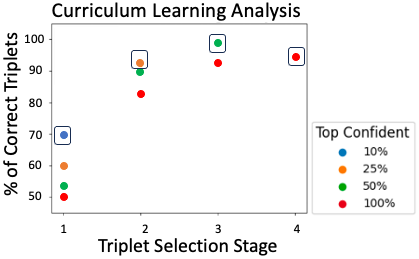}}
\caption{Curriculum Learning Analysis (as described in Fig.~\ref{fig:curriculum_learning_est}), using the ground-truth views instead of estimated views. The trend remains, but the percentage of correct triplets is higher.}
\label{fig:curriculum_learning_annotated}
\end{figure}

\textbf{Curriculum Learning.} To retrospectively analyze the effect of curriculum learning (Section~\ref{sec:curriculum_learning}), we evaluate the rate of correct triplets by determining triplets as \textit{correct} if $l_{a}$ = $l_{p}$ $\neq$ $l_{n}$ (based on true labels). Fig.~\ref{fig:curriculum_learning_est} shows that in each stage of Triplet Selection, selecting a smaller number of triplets with the highest confidence scores is better in terms of triplets' correctness. Moreover, the percentage of correct triplets increases with stages, as the model performance on the target domain improves. Both insights emphasize the importance of learning in an easy-to-hard manner. Although the percentage of correct triplets is approximately 15\%-20\% (Fig.~\ref{fig:curriculum_learning_est}), GOUDA significantly improves direct testing results, due to its ability to reduce the target domain view bias. Higher percentage (up to 90\%) is achieved when applying GOUDA using ground-truth views (see Fig.~\ref{fig:curriculum_learning_annotated}), highlighting the importance of quality view estimations.

\begin{figure}
\centerline{\includegraphics[width=0.55\linewidth]{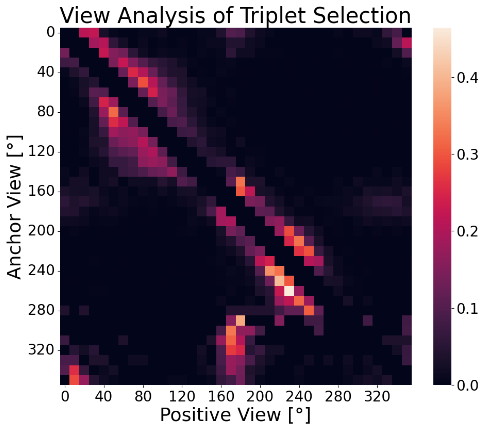}}
\caption{View analysis of Triplet Selection algorithm, presented on GaitGL from GREW (source) to OU-MVLP (target). Rows represent the anchor view and columns the selected positive view. Each row sums up to 1, and the measure presented is the percentage of each view of the selected positive samples. We suffer from a lack of diversity, due to the similarity of views. For example, for anchors with view 260, we mostly choose positives within the view range [210, 240].}
\label{fig:Triplets_analysis}
\end{figure}

\begin{figure}
\centerline{\includegraphics[width=\linewidth]{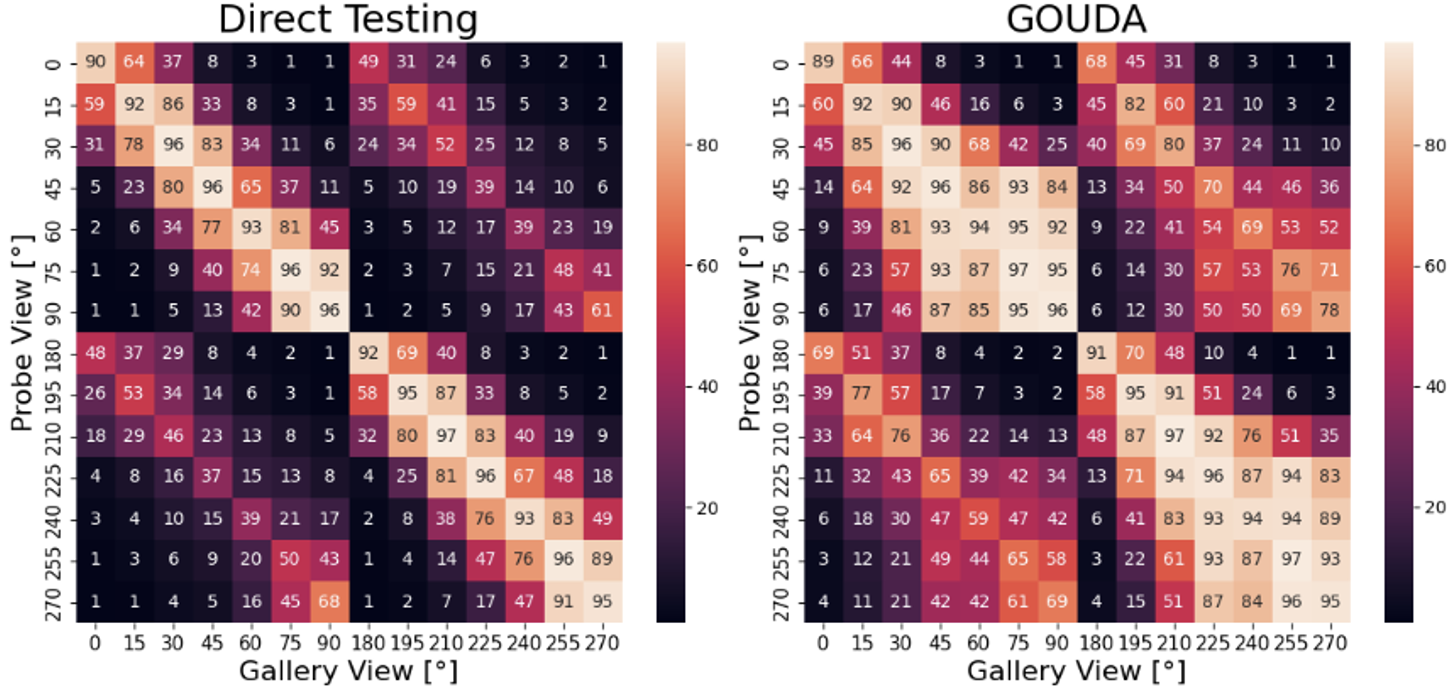}}
\caption{The internal Rank-1 accuracies on the target domain (OU-MVLP) test set. ``Direct Testing" is GaitGL pretrained in a supervised manner on the source domain (GREW). Each element in the matrix is the Rank-1 accuracy for the setting in which the probe view is $\alpha$ and the gallery set includes sequences of view $\beta$. Brighter color represents higher Rank-1. The final Rank-1 is the average of all Rank-1 accuracies such that $\alpha \neq \beta$. We present improvements in most of these cases after applying GOUDA.}
\label{fig:oumvlp_test_r1_analysis}
\end{figure}

\textbf{View Analysis.} Here we show the recognition accuracy as a function of view. Let us consider a triplet $\{a, p, n\}$ from our triplet selection algorithm. As described in Section~\ref{sec:triplet_selection}, $p$ is the closest sample to $a$ out of all \textit{cross view} samples. This results in a limited diversity among the selected triplets, as each view is closely associated with a specific set of views based on its gait features (see Fig.~\ref{fig:Triplets_analysis}). Despite that, we achieve significant improvements on the target domain test set. Fig.~\ref{fig:oumvlp_test_r1_analysis} presents the accuracy distribution on the test set according to view, before and after applying GOUDA. It is evident that the results improved for most of the cross view combinations, due to the indirect effect of changing the target domain view distribution (brighter colors in the right figure, implying higher accuracies). Furthermore, our method maintains the initial direct testing results on the exact same view combinations (diagonal), avoiding any degradation effects.

\section{Summary}
\label{sec:conclusion}
In this paper, we introduce a core problem of Gait Recognition models, often exhibiting a strong bias toward viewing angles when transferred to a new target domain. To reduce the presented gap causing a significant performance drop, we propose GOUDA - an unsupervised domain adaptation method, based on a new triplet selection strategy. We evaluate our method on multiple gait backbones and datasets, showing significant improvements and demonstrating superiority over prior works. We view our work as a first step toward exploring approaches to adapt gait models to target domains by looking at their view distributions.

{\small
\bibliographystyle{ieee_fullname}
\bibliography{GaitUDABib}
}

\appendix


%

\crefname{section}{Sec.}{Secs.}
\Crefname{section}{Section}{Sections}
\Crefname{table}{Table}{Tables}
\crefname{table}{Tab.}{Tabs.}


{\Large{\textbf{Supplementary Materials}}}


\section{Evaluation}
\subsection{GOUDA with view annotations}
In the paper, we present GOUDA results on various target datasets. We provide the results of the entire pipeline, including using estimated views extracted from the View Extraction module. However, there are two indoor datasets, CASIA-B and OU-MVLP, that provide ground-truth view annotations in advance. In order to reduce the noise produced by the View Extraction module and purely assess our Triplet Selection algorithm, we present GOUDA results using the view annotations (see Table~\ref{table:GOUDAMainTabSupp}). It is apparent that the Rank-1 improvements are higher, in this case, indicating the potential of GOUDA if using more accurate views. Particularly, the results on OU-MVLP dataset show a significant change. As an example, there is a substantial increase in Rank-1, from 27.9\% with estimated views to 40.5\% with annotated views, when adapting GaitSet backbone from CASIA-B (source) to OU-MVLP (target). 

\begin{figure*}[t!]
    \centering
    \begin{subfigure}[t]{0.3\textwidth}
        \centering
        \includegraphics[height=1.6in]{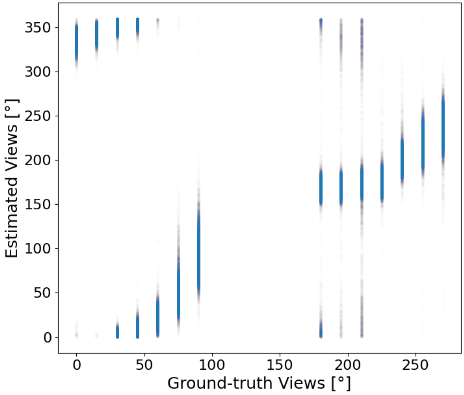}
        \caption{OU-MVLP}
    \label{fig:view_estimation_OUMVLP}
    \end{subfigure}
    ~ 
    \begin{subfigure}[t]{0.3\textwidth}
        \centering
        \includegraphics[height=1.6in]{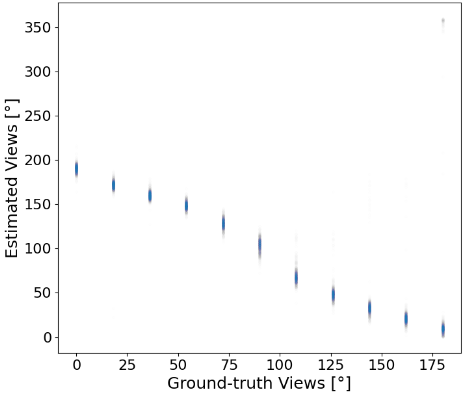}
        \caption{CASIA-B}
    \label{fig:view_estimation_CASIA}
    \end{subfigure}
    \caption{View estimation of OU-MVLP~\cite{takemura2018multi} and CASIA-B~\cite{yu2006framework} datasets, compared to their ground-truth view annotations provided as meta-data. The view estimation of OU-MVLP is limited due to the inaccurate 2D keypoints provided. Contrary to OU-MVLP, CASIA-B dataset includes RGB images that can be used to estimate accurate 2D keypoints, and therefore its view estimation is better (ambiguity of ±180 degrees is ignored).}
\label{fig:view_distribution_OU_CA}
\end{figure*}

In OU-MVLP dataset, RGB frames are not available (compared to CASIA-B dataset), and the view estimation is inferred from the quality of the provided 2D keypoints (Fig.~\ref{fig:view_estimation_OUMVLP}), especially compared to the view estimation of CASIA-B (Fig.~\ref{fig:view_estimation_CASIA}). These estimation errors might affect the Triplet Selection algorithm. The positive samples, which belong to the \textit{cross view} set, should indeed have different views than the anchor. However, if their views were estimated incorrectly, they might have entered to this set by mistake, and should have actually belong to the \textit{similar view} set. This unintended situation results in bringing closer samples with similar views, causing to not fulfilling the entire potential of GOUDA.

\begin{table*}
\renewcommand{\arraystretch}{1.1}
  \begin{center}
    {\small{
\begin{tabular}{ M{2cm}| M{1.5cm} | M{2.4cm} M{2.4cm} M{2.4cm} | M{2.4cm} }
\toprule[1.2pt]
\multirow{3}{*}{Source Dataset} & \multirow{3}{*}{Backbone} & \multicolumn{4}{c}{Target Dataset} \\
\cline{3-6}
& & \multicolumn{3}{c|}{CASIA-B} & \multirow{2}{*}{OU-MVLP}  \\
& & NM & BG & CL & \\
\midrule[1.2pt]
\multirow{2}{*}{CASIA-B} & \shortstack{GaitSet \\ GaitPart  \\ GaitGL} &  & - &  & \shortstack{\textbf{40.5} - 27.9 - 9.6 \\ \textbf{34.9} - 27.5 - 10.8 \\ \textbf{41.4} - 34.0 - 16.2} \\
\midrule
\multirow{2}{*}{OU-MVLP} & \shortstack{GaitSet \\ GaitPart  \\ GaitGL} & \shortstack{\textbf{90.8} - 87.0 - 74.0 \\ \textbf{94.5} - 91.5 - 73.9 \\ \textbf{94.8} - 92.8 - 81.7} & \shortstack{\textbf{70.4} - 68.0 - 55.5 \\ \textbf{80.6} - 78.4 - 56.9 \\ \textbf{82.3} - 80.9 - 71.5} & \shortstack{\textbf{29.5} - 27.2 - 16.4 \\ 36.7 - \textbf{38.7} - 20.7 \\ 38.8 - \textbf{44.6} - 28.8} & - \\
\midrule
\multirow{2}{*}{GREW} & \shortstack{GaitSet \\ GaitPart  \\ GaitGL} & \shortstack{\textbf{82.1} - 76.9 - 65.6 \\ \textbf{85.7} - 81.4 - 69.2 \\ \textbf{88.9} - 82.7 - 69.8} & \shortstack{\textbf{61.6} - 56.8 - 44.9 \\ \textbf{70.1} - 68.3 - 52.1 \\ \textbf{79.3} - 61.1 - 73.2} & \shortstack{23.9 - \textbf{26.0} - 20.8 \\ 33.8 - \textbf{33.9} - 25.4 \\ \textbf{45.6} - 44.6 - 31.9} & \shortstack{\textbf{49.5} - 38.8 - 21.8 \\ \textbf{52.5} - 43.6 - 23.9 \\ \textbf{62.1} - 44.2 - 25.7} \\
\midrule
\multirow{2}{*}{Gait3D} & \shortstack{GaitSet \\ GaitPart  \\ GaitGL} & \shortstack{\textbf{80.2} - 73.4 - 62.8 \\ \textbf{80.1} - 76.7 - 61.8 \\ \textbf{82.3} - 79.0 - 63.5} & \shortstack{\textbf{58.1} - 51.9 - 45.8 \\ \textbf{65.4} - 61.1 - 47.0 \\ \textbf{70.3} - 65.4 - 51.2} & \shortstack{\textbf{21.7} - 20.4 - 11.9 \\ 23.0 - \textbf{24.5} - 13.7 \\ \textbf{29.6} - 24.9 - 16.3} & \shortstack{\textbf{44.5} - 41.9 - 20.8 \\ \textbf{37.4} - 32.2 - 19.1 \\ \textbf{40.4} - 37.0 - 23.7} \\
\bottomrule[1.2pt]
\end{tabular}
}}
\end{center}
\caption{\label{table:GOUDAMainTabSupp} Rank-1 accuracy of GOUDA on the target datasets CASIA-B and OU-MVLP~\cite{takemura2018multi, yu2006framework}, compared to direct testing results on the backbones~\cite{chao2019gaitset},~\cite{fan2020gaitpart},~\cite{lin2021gait}. Here, we present two different setups of GOUDA results, one with estimated views (as presented in the paper), and the other by using ground-truth annotated views. The results of GOUDA with view annotations are reported on the left, GOUDA results with estimated views are in the middle, and the direct testing results are on the right, separated by dashes. In most cases, the best results are achieved by using GOUDA with annotated views, avoiding any noise produced by wrong view estimations or provided 2D keypoints.}
\end{table*}

\begin{figure*}[t!]
\centerline{\includegraphics[width=0.3\linewidth]{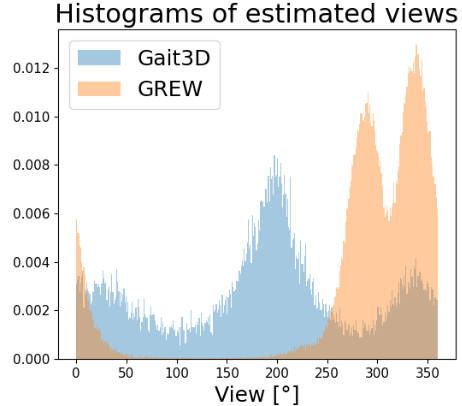}}
\caption{Histograms of estimated views for Gait3D and GREW datasets~\cite{Zheng_2022_CVPR, Zhu_2021_ICCV} using the View Extraction module.}
\label{fig:view_estimation_G3D_GR}
\end{figure*}

\begin{figure*}[t!]
\centerline{\includegraphics[width=\linewidth]{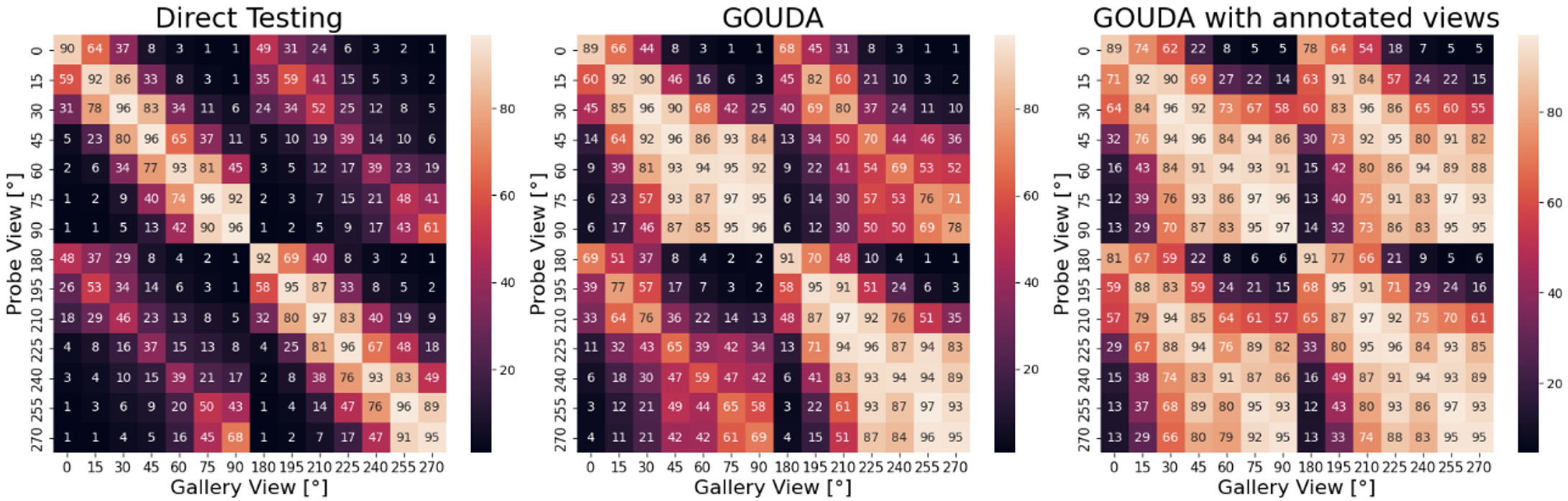}}
\caption{The internal Rank-1 accuracies on the target domain (OU-MVLP) test set. ``Direct Testing" is a gait backbone pretrained on the source domain (GREW). Each element in the matrix is the Rank-1 accuracy for the setting in which the probe view is $\alpha$ and the gallery set includes sequences of view $\beta$. Brighter color represents higher Rank-1. The final Rank-1 is the average of all Rank-1 accuracies such that $\alpha \neq \beta$. We present higher improvements when using view annotations (on the right) rather than using estimated views (in the middle).}
\label{fig:oumvlp_test_r1_analysis_supp}
\end{figure*}

\subsection{Applications} 
Suppose a real-world scenario of a target dataset with varied views but without true-positives (each person appears once). GOUDA can still be applied as it does not assume any conditions on the target labels. To demonstrate it, we create a subset of OU-MVLP dataset with a single appearance of each person and use it to train GOUDA using GaitGL pretrained on GREW. The Rank-1 accuracy improves from 25.7\% to 29.3\%. In this experiment we used OU-MVLP ground-truth view annotations.

\section{Analysis}
In this section, we provide additional analysis aspects, expanding the analysis presented in the paper. 

\subsection{GREW and Gait3D view estimation}
Both GREW~\cite{Zhu_2021_ICCV} and Gait3D~\cite {Zheng_2022_CVPR} were captured in the wild. Therefore, their view annotations are not provided. Fig.~\ref{fig:view_estimation_G3D_GR} presents the histograms of the estimated views for these datasets, using the proposed View Extraction module.

\subsection{View analysis}
Fig.~\ref{fig:oumvlp_test_r1_analysis_supp} presents the internal results on OU-MVLP test set by the end of GOUDA training. Two different cases are presented. In the first case, GOUDA was trained using OU-MVLP estimated views. In the second case, ground-truth view annotations were utilized for training. As detailed, the view estimation quality of OU-MVLP dataset is insufficient. Consequently, we observe greater improvements in internal Rank-1 accuracies when utilizing view annotations.

\subsection{Target view distribution}
\begin{figure*}[t!]
    \centering
    \begin{subfigure}[t]{0.237\textwidth}
        \centering
        \includegraphics[height=1.45in]{figures/umaps_sup/GaitGL_GR_OU/source.png}
        \caption{Target = Source}
    \label{fig:view_distribution_GL_source}
    \end{subfigure}
    ~ 
    \begin{subfigure}[t]{0.237\textwidth}
        \centering
        \includegraphics[height=1.45in]{figures/umaps_sup/GaitGL_GR_OU/cross_domain.png}
        \caption{Direct Testing}
    \label{fig:view_distribution_GL_cross}
    \end{subfigure}
    ~ 
    \begin{subfigure}[t]{0.237\textwidth}
        \centering
        \includegraphics[height=1.45in]{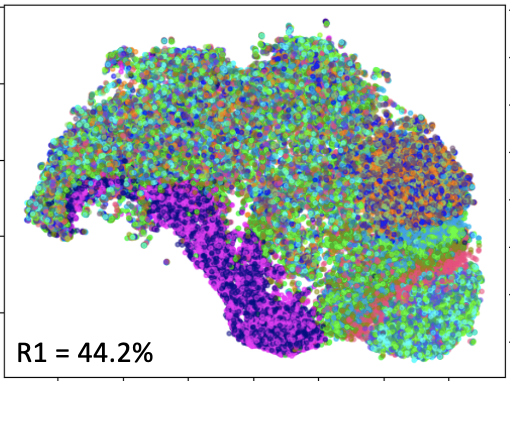}
        \caption{Direct Testing + GOUDA with estimated views}
    \label{fig:view_distribution_GL_gouda_est}
    \end{subfigure}
    ~ 
    \begin{subfigure}[t]{0.237\textwidth}
        \centering
        \includegraphics[height=1.45in]{figures/umaps_sup/GaitGL_GR_OU/gouda_annot.png}
        \caption{Direct Testing + GOUDA with annotated views}
    \label{fig:view_distribution_GL_gouda_annot}
    \end{subfigure}
    \caption{Gait embedding visualization (UMAP~\cite{mcinnes2018umap}) for domain transfer between GREW as source and OU-MVLP as the target domain using GaitGL. Points are color coded by viewing angle. Fig.~\ref{fig:view_distribution_GL_source} presents the single domain scenario, in which the model was trained on OU-MVLP training set.  Fig.~\ref{fig:view_distribution_GL_cross} presents the Direct Testing scenario, in which the model was trained on a different gait dataset (GREW). Fig.~\ref{fig:view_distribution_GL_gouda_est} presents the Direct Testing scenario after applying GOUDA using estimated views. Fig.~\ref{fig:view_distribution_GL_gouda_annot} presents the Direct Testing scenario after applying GOUDA using the ground-truth annotated views. GOUDA achieves higher improvements when using annotated views.}
\label{fig:view_distribution_GL}
\end{figure*}

\begin{figure*}[t!]
    \centering
    \begin{subfigure}[t]{0.237\textwidth}
        \centering
        \includegraphics[height=1.45in]{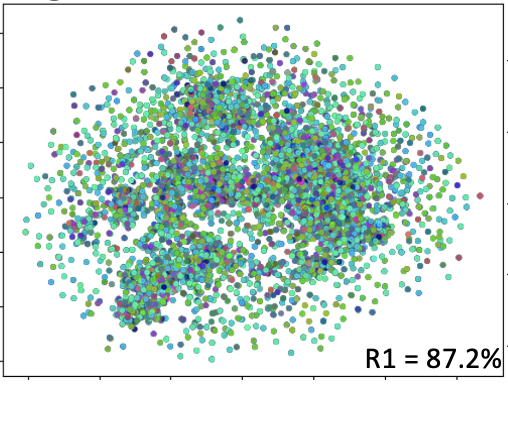}
        \caption{Target = Source}
    \label{fig:view_distribution_set_source}
    \end{subfigure}
    ~ 
    \begin{subfigure}[t]{0.237\textwidth}
        \centering
        \includegraphics[height=1.45in]{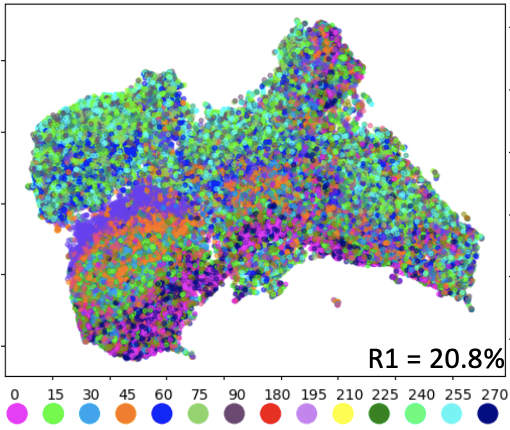}
        \caption{Direct Testing}
    \label{fig:view_distribution_set_cross}
    \end{subfigure}
    ~ 
    \begin{subfigure}[t]{0.237\textwidth}
        \centering
        \includegraphics[height=1.45in]{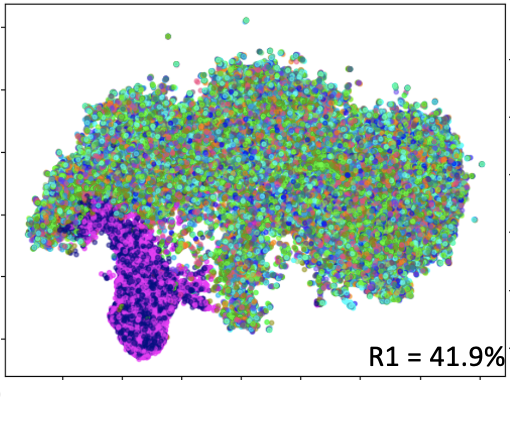}
        \caption{Direct Testing + GOUDA with estimated views}
    \label{fig:view_distribution_set_gouda_est}
    \end{subfigure}
    ~ 
    \begin{subfigure}[t]{0.237\textwidth}
        \centering
        \includegraphics[height=1.45in]{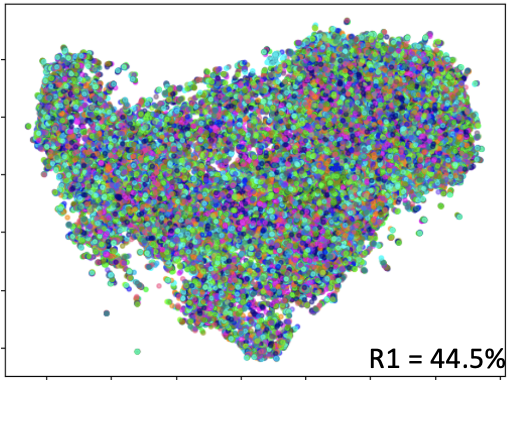}
        \caption{Direct Testing + GOUDA with annotated views}
    \label{fig:view_distribution_set_gouda_annot}
    \end{subfigure}
    \caption{Gait embedding visualization (UMAP~\cite{mcinnes2018umap}) for domain transfer between Gait3D as source and OU-MVLP as the target domain using GaitSet. Points are color coded by viewing angle. Fig.~\ref{fig:view_distribution_set_source} presents the single domain scenario, in which the model was trained on OU-MVLP training set.  Fig.~\ref{fig:view_distribution_set_cross} presents the Direct Testing scenario, in which the model was trained on a different gait dataset (Gait3D). Fig.~\ref{fig:view_distribution_set_gouda_est} presents the Direct Testing scenario after applying GOUDA using estimated views. Fig.~\ref{fig:view_distribution_set_gouda_annot} presents the Direct Testing scenario after applying GOUDA using the ground-truth annotated views. GOUDA achieves higher improvements when using annotated views.}
\label{fig:view_distribution_set}
\end{figure*}

\begin{figure*}[t!]
    \centering
    \begin{subfigure}[t]{0.237\textwidth}
        \centering
        \includegraphics[height=1.45in]{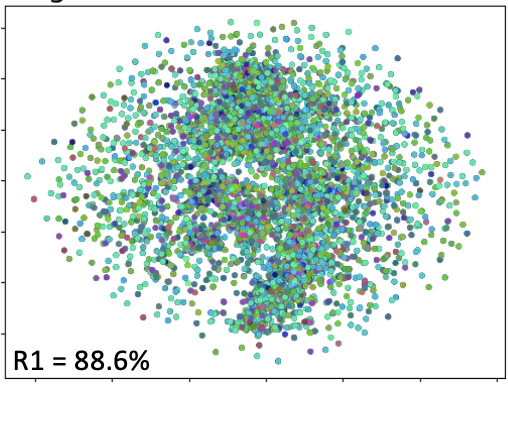}
        \caption{Target = Source}
    \label{fig:view_distribution_Part_source}
    \end{subfigure}
    ~ 
    \begin{subfigure}[t]{0.237\textwidth}
        \centering
        \includegraphics[height=1.45in]{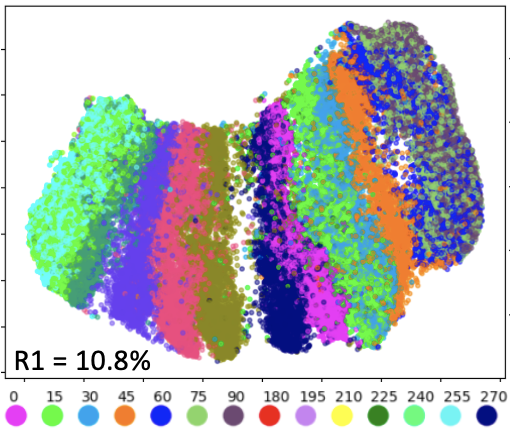}
        \caption{Direct Testing}
    \label{fig:view_distribution_Part_cross}
    \end{subfigure}
    ~ 
    \begin{subfigure}[t]{0.237\textwidth}
        \centering
        \includegraphics[height=1.45in]{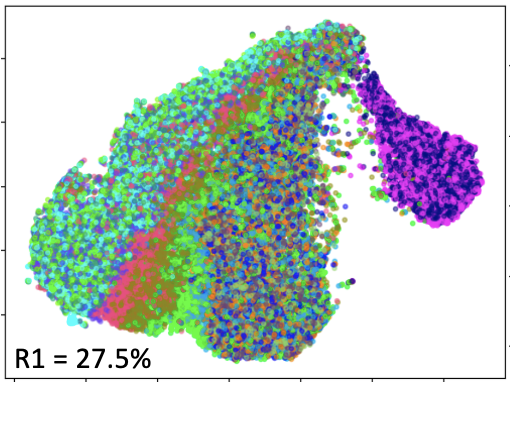}
        \caption{Direct Testing + GOUDA with estimated views}
    \label{fig:view_distribution_Part_gouda_est}
    \end{subfigure}
    ~ 
    \begin{subfigure}[t]{0.237\textwidth}
        \centering
        \includegraphics[height=1.45in]{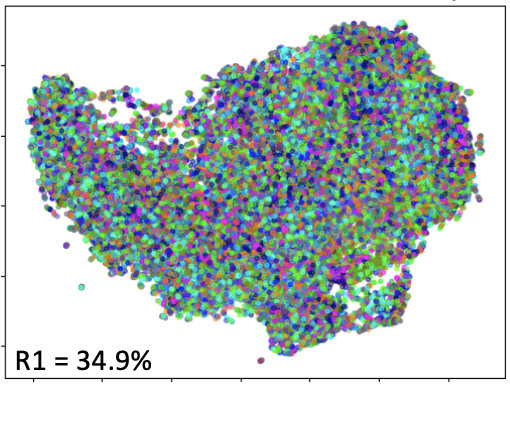}
        \caption{Direct Testing + GOUDA with annotated views}
    \label{fig:view_distribution_Part_gouda_annot}
    \end{subfigure}
    \caption{Gait embedding visualization (UMAP~\cite{mcinnes2018umap}) for domain transfer between CASIA-B as source and OU-MVLP as the target domain using GaitPart. Points are color coded by viewing angle. Fig.~\ref{fig:view_distribution_Part_source} presents the single domain scenario, in which the model was trained on OU-MVLP training set.  Fig.~\ref{fig:view_distribution_Part_cross} presents the Direct Testing scenario, in which the model was trained on a different gait dataset (CASIA-B). Fig.~\ref{fig:view_distribution_Part_gouda_est} presents the Direct Testing scenario after applying GOUDA using estimated views. Fig.~\ref{fig:view_distribution_Part_gouda_annot} presents the Direct Testing scenario after applying GOUDA using the ground-truth annotated views. GOUDA achieves higher improvements when using annotated views.\\}
\label{fig:view_distribution_Part}
\end{figure*}

In the paper, we describe a fundamental phenomenon in gait recognition models. As detailed, gait recognition models place a strong emphasis on view-based features in the target domain, exposing the huge gap between source and target domains. The observed behavior is not unique to any particular gait backbone, and is consistent across different backbones that we checked. In this context, we present similar patterns for additional gait recognition models beyond what is covered in the paper (see Figs.~\ref{fig:view_distribution_GL}, ~\ref{fig:view_distribution_set}, and~\ref{fig:view_distribution_Part}). The pattern is less clear (but still exists) when using GaitSet as the backbone (Fig.~\ref{fig:view_distribution_set}), probably because it perceives the silhouettes as a set of images rather than a temporal sequence, thus assigning less significance to the view information.

\subsection{Curriculum Learning}
Further ablation experiments on the curriculum learning hyperparameter $q$ (the percentage of top confident valid triplets selected for training in each stage) are reported in Table~\ref{table:AblationStudySupp}. These results are inferior or comparable to those achieved with the original $q$ values: 10, 25, 50, and 100.

\begin{table}
\renewcommand{\arraystretch}{1.1}
  \centering
  \small
  \begin{tabular}{M{3.7cm}| M{0.8cm} | M{0.8cm} | M{0.8cm}}
    \toprule[1.2pt]
    \multirow{2}{*}{$q$ values [\%]} & \multicolumn{3}{c}{OU-MVLP $\rightarrow$ CASIA-B} \\
    \cline{2-4}
    & NM & BG & CL \\
    \midrule[1pt]
    5, 25, 70, 100  & 91.3 & 76.7 & 39.7 \\
    \hline
    10, 20, 40, 100  & 92.4 & 80.3 & 47.2 \\
    \hline
    15, 30, 60, 100  & 90.1 & 74.1 & 40.6 \\
    \hline
    20, 50, 75, 100  & 91.5 & 76.1 & 38.1 \\  
    \hline
    \textbf{GOUDA} (10, 25, 50, 100) & 92.8 & 80.9 & 44.6 \\
    \bottomrule[1.2pt]
  \end{tabular}

\caption{\label{table:AblationStudySupp}
Ablation experiments on the curriculum learning hyperparameter $q$. In this setting, OU-MVLP is the source domain and CASIA-B is the target domain. Average Rank-1 accuracies across all 11 views, are shown, excluding identical-view cases.}
\end{table}

\section{Justification}

\subsection{Positive sample selection}
\begin{figure*}[t!]
    \centering
    \begin{subfigure}[t]{0.4\textwidth}
        \centering
        \includegraphics[height=2.3in]{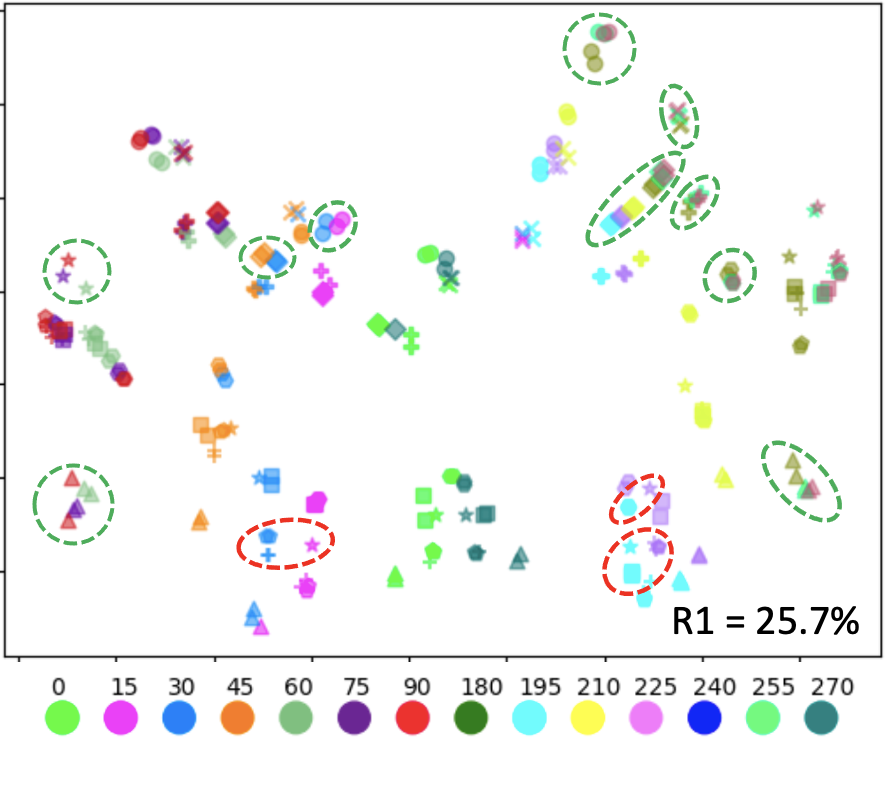}
        \caption{Direct Testing}
    \label{fig:justification_direct_testing}
    \end{subfigure}
    ~ 
    \begin{subfigure}[t]{0.4\textwidth}
        \centering
        \includegraphics[height=2.3in]{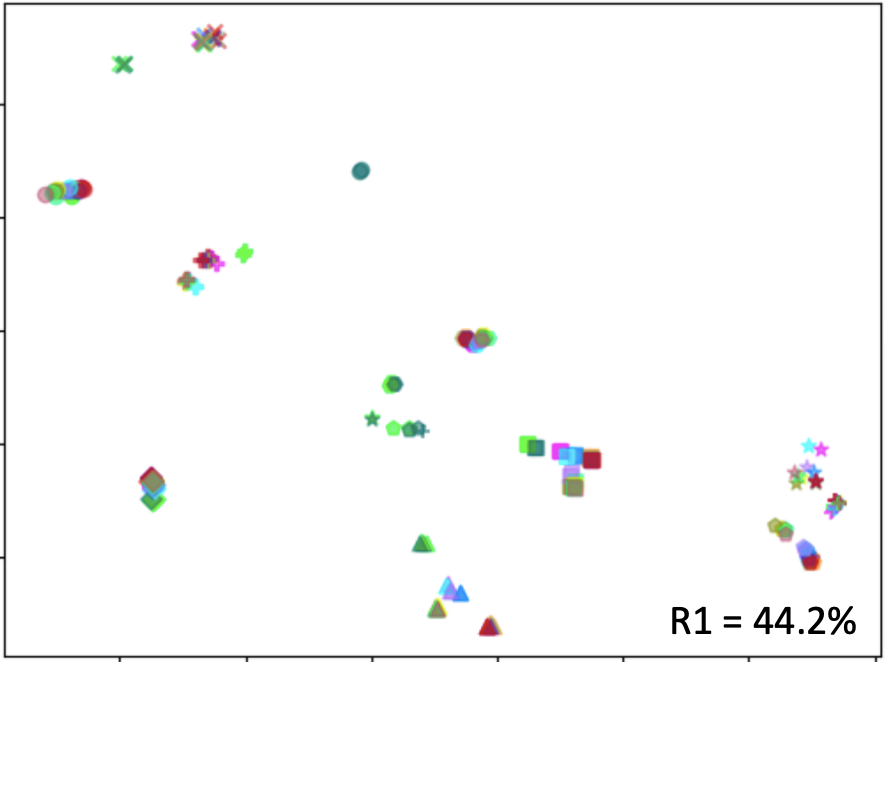}
        \caption{Direct Testing + GOUDA}
    \label{fig:justification_gouda_est}
    \end{subfigure}
    \caption{Gait embedding visualization (UMAP~\cite{mcinnes2018umap}) for domain transfer between GREW as source and OU-MVLP as the target domain using GaitGL. Points are color coded by viewing angle while identities are discriminated via the plot shapes. For the sake of visualization, only $10$ identities are shown. Fig.~\ref{fig:justification_direct_testing} presents the Direct Testing scenario, in which the model was trained on a different gait dataset (GREW). The green/red dashed curves present the case in which the closest sample with different view shares the same/different identity, respectively. Fig.~\ref{fig:justification_gouda_est} presents the distribution after applying GOUDA. Same identities are well clustered, reflected by higher Rank-1 accuracy.}
\label{fig:justification}
\end{figure*}

Here, we illustrate the intuition behind our positive sample strategy. Fig.~\ref{fig:justification} presents a shared visualization of viewing angles (color coded) and identities (shapes) on a subset of the OU-MVLP target dataset. For the sake of visualization, we present only $10$ identities. In Fig.~\ref{fig:justification_direct_testing}, the green dashed curves illustrate the cases in which our hypothesis is satisfied, meaning that the closest samples with different views share the same identity. Conversely, the red-bordered regions showcase ``optional" breach of our hypothesis. Our approach effectively functions under ``noisy" pseudo labeled data. To this end, our curriculum learning protocol is structured to potentially avoid the choice of anchors from the red-bordered areas. This gradual model enhancement, as advocated by curriculum learning, consequently fosters the selection of anchors from the green areas while contributing to the reduction of red-bordered regions by gradual target adaption. This result is justified by the ramp-up in cross-view R1 accuracy after the adaptation (from 25.7\% to 44.2\%), and with better identity-based clustering, as demonstrated in Fig.~\ref{fig:justification_gouda_est}.

\begin{figure}
\centerline{\includegraphics[width=0.7\linewidth]{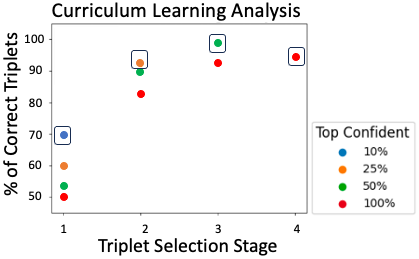}}
\caption{Curriculum Learning Analysis. Triplet Selection is applied $4$ times during training (x-axis). Y-axis shows how many of the selected triplets are correct considering person identities. The colors represent the top confident triplets, and in each stage the chosen percentage is marked by a rectangle. Here, GaitGL is used for domain transfer from GREW (source) to OU-MVLP (target) with the ground-truth annotated views.}
\label{fig:curriculum_learning_annotated_supp}
\end{figure}

\subsection{Quantitative evaluation}
To further support the triplet selection strategy, we present Table~\ref{table:justification}. It depicts the percentage of selecting the \textit{correct} triplets, as well as the correct positive samples and negative samples, separately. The positive sample is correct if it shares the same identity as the anchor, whereas the negative sample is correct if it has a different identity. We show two different settings using the initial pre-trained source model (first curriculum learning phase). It is shown that the selection percentage of both positive and negative samples is substantial, even under ``noisy" conditions. Moreover, by employing the curriculum learning approach, the model performance on the target domain is improved throughout training, leading to an increase in the proportion of correct triplets (see Fig.~\ref{fig:curriculum_learning_annotated_supp}).

\begin{table}
\renewcommand{\arraystretch}{1.1}
  \centering
  \small
  \begin{tabular}{M{3.0cm}| M{1.15cm} | M{1.15cm} | M{1.15cm}}
    \toprule[1.2pt]
    {Target Dataset}
    & Triplets & Positives & Negatives \\
    \midrule[1pt]
    CASIA-B & 88.1 & 98.5 & 88.9 \\
    \hline
    OU-MVLP & 69.4 & 69.4 & 100 \\
    \bottomrule[1.2pt]
  \end{tabular}
\caption{\label{table:justification}
Identity-wise correctness [\%] of the selected triplets in the first curriculum learning phase. The correctness of the positive and negative samples is presented as well. A positive sample is correct if it shares the same anchor identity, while a negative sample is correct if it has a different identity. Here, GaitGL is used for domain transfer from GREW (source) to CASIA-B or OU-MVLP (target) using the ground-truth annotated views.}
\end{table}


\end{document}